\documentclass[conference]{IEEEtran} 
\IEEEoverridecommandlockouts
\usepackage{todonotes}
\usepackage{cite}
\usepackage{amsmath,amssymb,amsfonts}
\usepackage{algorithmic}
\usepackage{graphicx}
\usepackage{textcomp}
\usepackage{xcolor}
\def\BibTeX{{\rm B\kern-.05em{\sc i\kern-.025em b}\kern-.08em
    T\kern-.1667em\lower.7ex\hbox{E}\kern-.125emX}}
\begin{document}

\title{Enhancing Knowledge Graph Construction Using  Large Language Models}

\author{\IEEEauthorblockN{1\textsuperscript{st} Milena Trajanoska}
\IEEEauthorblockA{\textit{Faculty of Comp. Sci. and Eng.} \\
\textit{Ss. Cyril and Methodius University}\\
Skopje, Macedonia \\
milena.trajanoska@finki.ukim.mk\\
ORCID: 0000-0003-0105-7693}
\and
\IEEEauthorblockN{2\textsuperscript{nd} Riste Stojanov}
\IEEEauthorblockA{\textit{Faculty of Comp. Sci. and Eng.} \\
\textit{Ss. Cyril and Methodius University}\\
Skopje, Macedonia \\
riste.stojanov@finki.ukim.mk\\
ORCID: 0000-0003-2067-3467}
\and
\IEEEauthorblockN{3\textsuperscript{rd} Dimitar Trajanov}
\IEEEauthorblockA{\textit{Faculty of Comp. Sci. and Eng.} \\
\textit{Ss. Cyril and Methodius University}\\
Skopje, Macedonia \\
dimitar.trajanov@finki.ukim.mk\\
ORCID: 0000-0002-3105-6010}
}

\maketitle

\begin{abstract}
The growing trend of Large Language Models (LLM) development has attracted significant attention, with models for various applications emerging consistently. However, the combined application of Large Language Models with semantic technologies for reasoning and inference is still a challenging task. This paper analyzes how the current advances in foundational LLM, like ChatGPT, can be compared with the specialized pretrained models, like REBEL, for joint entity and relation extraction. 
To evaluate this approach, we conducted several experiments using sustainability-related text as our use case. 
We created pipelines for the automatic creation of Knowledge Graphs from raw texts, and our findings indicate that using advanced LLM models can improve the accuracy of the process of creating these graphs from unstructured text. 
Furthermore, we explored the potential of automatic ontology creation using foundation LLM models, which resulted in even more relevant and accurate knowledge graphs.

\end{abstract}


\section{Introduction}
The technological advancements, together with the availability of Big Data, have led to a surge in the development of Large Language Models (LLMs) \cite{brants2007large}. This trend has paved the way for a cascade of new models being released on a regular basis, each outperforming its predecessors. These models have started a revolution in the field with their capability to process massive amounts of unstructured text data and by achieving state-of-the-art results on multiple Natural Language Processing (NLP) tasks. 

However, one of the aspects which have not yet taken over the spotlight is the combined application of these models with semantic technologies to enable reasoning and inference. This paper attempts to fill this gap by making a connection between the Deep Learning (DL) space and the semantic space, through the use of NLP for creating Knowledge Graphs \cite{chen2020review}. 

Knowledge Graphs are structured representations of information that capture the relationships between entities in a particular domain. They are used extensively in various applications, such as search engines, recommendation systems, and question-answering systems.

On a related note, there is a significant amount of raw texts available on the Web which contain valuable information. Nevertheless, this information is unusable if it cannot be extracted from the texts and applied for intelligent reasoning. This fact has motivated us to use some of the state-of-the-art models in an attempt to extract information from text data on the Web.

Yet, creating Knowledge Graphs from raw text data is a complex task that requires advanced NLP techniques such as Named Entity Recognition \cite{mikheev1999named}, Relation Extraction \cite{zhou2005exploring}, and Semantic Parsing \cite{kamath2018survey}. Large language models such as GPT-3 \cite{brown2020language}, T5 \cite{raffel2020exploring}, and BERT \cite{devlin2018bert} have shown remarkable performance in these tasks, and their use has resulted in significant improvements in the quality and accuracy of knowledge graphs.

To evaluate our approach in connecting both fields, we chose to analyze the specific use case of sustainability. Sustainability is a topic of great importance for our future, and a lot of emphasis has been placed on identifying ways to create more sustainable practices in organizations. Sustainability has become the norm for organizations in developed countries, mainly due to the rising awareness of their consumers and employees. However, this situation is not reflected in developing and underdeveloped countries to this extent. Although the perception of sustainability has improved, progress toward sustainable development has been slower, indicating the need for more concrete guidance \cite{nations_2018}. Moreover, theoretical research has attempted to link strategic management and sustainable development in corporations in order to encourage the integration of sustainability issues into corporate activities and strategies \cite{baumgartner2017strategic}. Even though research has set a basis for developing standards and policies in favor of sustainability, a more empirical approach is needed for policy definitions and analyzing an organization's sustainability level with respect to the defined policies.

In this study, the goal is to make a connection between LLMs and semantic reasoning to automatically generate a Knowledge Graph on the topic of sustainability and populate it with concrete instances using news articles available on the Web. For this purpose, we create multiple experiments where we utilize popular NLP models, namely Relation Extraction By End-to-end Language generation (REBEL) \cite{rebel} and ChatGPT \cite{openai_gpt-4_2023}. We show that although REBEL is specifically trained for relation extraction, ChatGPT, a conversational agent using a generative model, can streamline the process of automatically creating accurate Knowledge Graphs from an unstructured text when provided with detailed instructions.

The rest of the paper is structured as follows: Section II presents a brief literature overview, Section III describes the methods and experimental setup, Section IV outlines the results of the information extraction process, Section V states the propositions for future work, and finally section VI gives the conclusion of the work done in this paper.

\section{Literature Review}

\subsection{Algorithms}

Our study focuses on the task of information extraction from news and reports available on the Web. For this purpose, we compare the capabilities of NLP models to generate a useful Knowledge Base on the topic.


A Knowledge Base represents information stored in a structured format, ready to be used for analysis or inference. Often, Knowledge Bases are stored in the form of a graph and are then called Knowledge Graphs.

In order to create such a Knowledge Base, we need to extract information from the raw texts in a triplet format. An example of a triplet would be $<$Person, Location, City$>$. In the triplet, we have a structure consisting of the following links Entity -$>$ Relation -$>$ Entity, where the first entity is referred to as the subject, the relation is a predicate, and the second entity represents the object. In order to achieve this structured information extraction, we need to identify entities in the raw texts, as well as the relations connecting these entities. 

In the past, this process was implemented by leveraging multi-step pipelines, where one step included Named-entity Recognition (NER) \cite{mikheev1999named}, and another step was Relation classification (RC) \cite{zeng2014relation}. However, these multi-step pipelines often prove to have unsatisfactory performance due to the propagation of errors from the steps. In order to tackle this problem, end-to-end approaches have been implemented, referred to as Relation-Extraction (RE) \cite{zhou2005exploring} methods.


One of the models utilized in this study is REBEL (Relation Extraction By End-to-end Language generation) \cite{rebel}, which is an auto-regressive seq2seq model based on BART \cite{lewis-etal-2020-bart} that performs end-to-end relation extraction for more than 200 different relation types. The model achieves 74 micro-F1 and 51 macro-F1 scores. It was created for the purpose of joint entity-relation extraction.

REBEL is a generative seq2seq model which attempts to "translate" the raw text into a triple format. The REBEL model outputs additional tokens, which are used during its training to identify a triplet. These tokens include $<$triplet$>$, which represents the beginning of a triplet, $<$subj$>$, which represents the end of the subject and the start of the predicate, and $<$obj$>$, which represents the end of the predicate and start of the object. The authors of the paper for REBEL provide a parsing function for extracting the triplet from the output of REBEL.

The second approach we took was to use ChatGPT \cite{openai_gpt-4_2023}, as a 
conversational agent and compare the performance in the task of entity-relation extraction and creation of a common Knowledge Base. The agent consists of three steps, including separate models: a supervised fine-tuning (SFT) model based on GPT-3 \cite{brown2020language}, a reward model, and a reinforcement learning model. 

 ChatGPT was trained using Reinforcement Learning from Human Feedback (RLHF) \cite{ouyang2022training}, employing methods similar to InstructGPT with minor variations in data collection. An initial model is trained through supervised fine-tuning, with human AI trainers engaging in conversations, assuming both user and AI assistant roles. To aid in formulating responses, trainers were given access to model-generated suggestions. The newly created dialogue dataset was then combined with the InstructGPT dataset, which was transformed into a dialogue format. In order to establish a reward model for reinforcement learning, comparison data needed to be gathered, consisting of two or more model responses ranked by quality. This data was collected by taking conversations between AI trainers and the chatbot, randomly selecting a model-generated message, sampling multiple alternative completions, and having AI trainers rank them. The reward models enabled fine-tuning of ChatGPT using Proximal Policy Optimization \cite{schulman2017proximal}, and several iterations of this procedure were executed.

\subsection{Use case: Sustainability}
The Global sustainability study of 2022 has reported that 71\% out of 11,500 surveyed consumers around the world are making changes to the way they live and the products they buy in an effort to live more sustainably \cite{global-sustainability}. This shows that corporations not only need to change their operations to be more sustainable for the sake of the environment but also to be able to stay competitive.

With the vast amount of  unstructured data available on the Web, it is crucial to develop methods that can automatically identify sustainability-related information from news, reports, papers, and other forms of documents. One such study identifies this opportunity and attempts to create a method for directly extracting non-financial information generated by various media to provide objective ESG information \cite{lee2023esg}. The authors have trained an ESG classifier and recorded a classification accuracy of 86.66\% on 4-class on texts which they manually labeled. On a related note, researchers have taken a step further to extract useful ESG information from texts. In this article \cite{ehrhardt2022automated}, the authors have trained a joint entity and relation extraction model on a private dataset consisting of ESG and CSR reports annotated internally at Crédit Agricole. They were able to identify entities such as coal activities and environmental or social issues. In \cite{vodenska2022challenges}, the authors presented an approach for knowledge graph generation based on ESG-related news and company official documents.

\section{Methods}

This section describes the methods used in this research, including the data collection process and the entity-relation extraction algorithms used to analyze the gathered data.

\subsection{Data Collecting Process}

In order to conduct the experimental comparison of the two approaches for entity-relation extraction, news data was gathered from the Web on the topic of sustainability. For this purpose, the News API \cite{newsapi} system was used. News API is an HTTP REST API for searching and retrieving live articles from all over the Web. It provides the ability to search through the articles posted on the Web by specifying the following options: keyword or phrase, date of publication, source domain name, and language.

Using News API, 94 news articles from 2023-02-15 to 2023-03-19 on the topic of sustainability have been collected. The collected texts contained various numbers of words ranging from 50 to over 4200. With the limitation of the number of tokens that can be passed as input to a language model, additional pre-processing steps needed to be taken to account for the texts consisting of a large number of words.

\subsection{Relation-Extraction Methods}

Relation-extraction is a fundamental task in NLP that aims to identify the semantic relationships between entities in a sentence or document. The task is challenging because it requires understanding the context in which the entities appear and the types of relationships that exist between them.

In this subsection, we describe how we utilize REBEL and ChatGPT for the task of relation extraction.

\subsubsection{REBEL}

Our first approach was to use REBEL in an attempt to extract relations from unstructured news articles. In order for REBEL to be able to use the provided texts, they need to be tokenized with the corresponding tokenizer function. Tokenization is the process of separating the raw text into smaller units called tokens. Tokens can refer to words, characters, or sub-words. The model has a token limitation of 512 tokens, which means that the collected articles which are longer need to be pre-processed before sending them to the model for triplets extraction. 

To address this limitation, we tokenize the raw text and divide the tokens into 256-token batches. These batches are processed separately by the REBEL model, and the results are subsequently merged to extract relations for longer texts. Metadata is also added to the extracted relations, referencing the token batch from which the relation was derived. With this approach, some relations may not be extracted accurately because the batch of tokens might begin or end in the middle of the sentence. However, the number of cases where this happens is insignificant. Thus, we leave their handling for future work.

Once the entity-relation extraction process is finished, the extracted information is stored in a triplet structure. To further normalize the extracted entities, we perform Entity Linking \cite{shen2014entity}. Entity Linking refers to the identification and association of entity mentions in raw text with their corresponding entities in a Knowledge Base. The process of Entity Linking is not part of the REBEL model, and it is an additional post-processing step that is used to refine the extracted relations. In this study, we utilize DBpedia as our Knowledge Base and consider two entities identical if they share the same DBpedia URL. This approach will not work for entities that are not present on DBpedia.

\subsubsection{ChatGPT}

The second approach taken in this paper uses OpenAI's ChatGPT \cite{openai_gpt-4_2023}. We have created two experiments using ChatGPT. 

The first experiment prompts ChatGPT to extract relations from the collected news articles. After extracting the relations, we follow the same steps as with the REBEL model in order to create a comprehensive Knowledge Base.

The second experiment focuses on creating a prompt that would directly generate the entire Knowledge Base and write an ontology describing the concepts identified in the texts. This approach has the goal of reducing the number of manual steps which need to be performed in order to obtain the final Knowledge Graph.

For both experiments, we set the value of the parameter 'temperature' to 0 in order to get more deterministic outputs since OpenAI models are non-deterministic by nature.

\textbf{Experiment 1}. For the first experiment, we prompt ChatGPT to extract relations connected to sustainability. ChatGPT was able to successfully extract entities and connect them with relations, and return the results in a triple format. After the relations had been extracted, the same post-processing step of Entity Linking was implemented on the results from ChatGPT.

Although ChatGPT was able to extract entities from the articles and link them with relations, it was not successful at abstracting concepts. The entities and relations identified often represented whole phrases instead of concepts.

To overcome the obstacle, we prompted ChatGPT to map identified entities and relations to a suitable OWL ontology \cite{mcguinness2004owl}. However, ChatGPT failed to identify relevant sustainability concepts or define their instances. The identified classes, such as \textit{Company}, \textit{Customer}, \textit{MarketingEcosystem}, \textit{Resource}, \textit{CustomerExperience}, \textit{Convenience}, and \textit{DigitalMarketing}, had some potential relevance to sustainability, but ChatGPT did not identify any instances for these classes.

\textbf{Experiment 2}. In the second experiment, we refined the prompt to ask ChatGPT to explicitly generate an OWL ontology on sustainability, which includes concepts like \textit{organizations, actions, practices, policies}, and related terms. We also allowed ChatGPT to create additional classes and properties if necessary. We explicitly requested the results to be returned in RDF Turtle format.

Providing additional information to ChatGPT resulted in the creation of an improved Knowledge Base. ChatGPT was able to define concepts such as \textit{organizations}, \textit{actions}, \textit{practices}, and \textit{policies}, as well as identify suitable relations to connect them together. Moreover, it was able to create instances of the defined classes and properties and link them together. This shows that adding more specific instructions to the prompts for ChatGPT can produce drastically different results.

\section{Results}
This section presents the results from the experiments described in Section III. A comparison of the created Knowledge Base from both methods is given, and the characteristics of the generated Knowledge Bases are outlined. Table \ref{tab:table1} represents the Knowledge Bases from the REBEL model and the first experiment with ChatGPT, respectively. The table shows the number of entities, relations, and triplets extracted from the raw texts on sustainability.

\begin{table}[htbp]
\caption{Knowledge Base structure comparison}
\label{tab:table1}
\centering
\begin{tabular}{|l|l|l|l|}
\hline
\textbf{Algorithm} & \textbf{Entities} & \textbf{Relations} & \textbf{Triples} \\ \hline
REBEL              & 805               & 105                & 854              \\ \hline
ChatGPT            & 1158              & 677                & 826              \\ \hline
\end{tabular}
\end{table}

As it is evident from the table, the number of triplets extracted by both algorithms is similar. However, the number of entities that ChatGPT extracts are larger than those from REBEL. Although this is true, a lot of the extracted entities are not connected to each other via any relation, thus defeating the purpose of creating a Knowledge Base. Moreover, the number of unique relations is far too large for ChatGPT to be able to produce an ontology that can be used for further experimentation. 

The most frequent relation for the REBEL model is the \textit{'subclass of'} relation, being part of \textit{120} triplets. For ChatGPT, it's the \textit{'has'} relation, being identified in \textit{29} triplets. In addition, ChatGPT often fails to generate standard relations and entities which represent abstract concepts and instead outputs an entire phrase, such as in the example 'has already surpassed a goal set in 2019 to install 100,000 heat pumps in homes and businesses', where it identifies this phrase as a relation.

The following subsections represent a visual display of a subset of the generated Knowledge Bases from both algorithms.

\subsection{REBEL}

\begin{figure}[htbp]
\label{fig:fig_rebel}
\centerline{\includegraphics[width=0.4\textwidth]{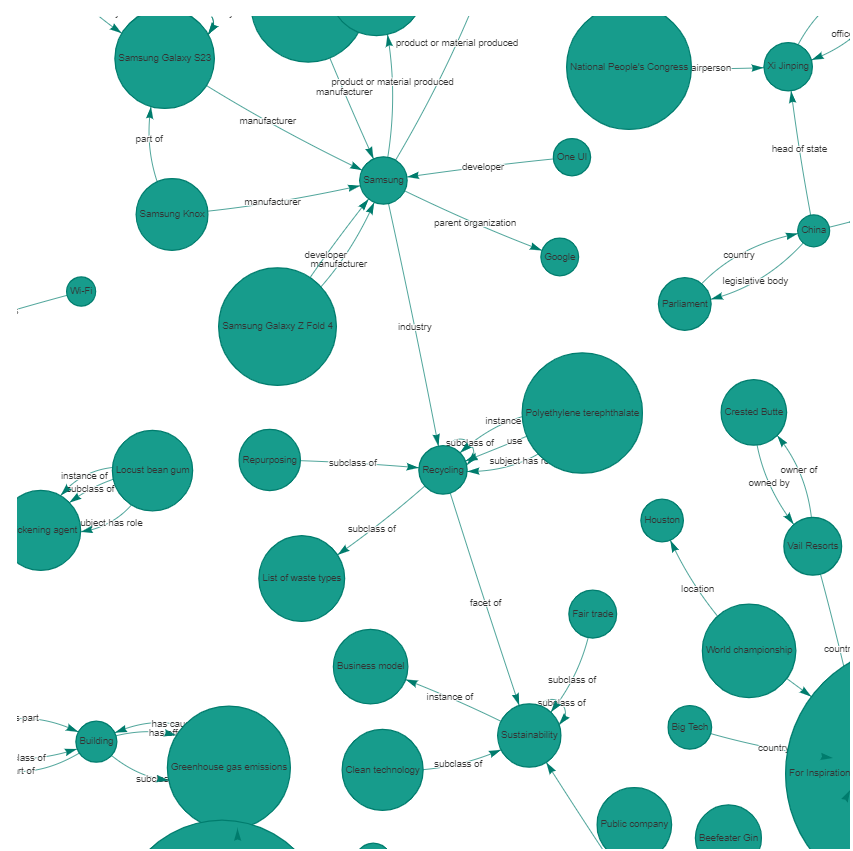}}
\caption{Subset of the Knowledge Base generated using the REBEL model. The Knowledge Base is displayed in a graph format where entities are represented as nodes and relations are represented as edges.}
\end{figure}

In order to be able to analyze the Knowledge Base generated using the REBEL model more accurately, we have created a visualization in a graph format, where each entity represents a node in the graph, and each relation represents an edge. Fig. \ref{fig:fig_rebel} displays a subset of the extracted Knowledge Base.

It is visible from the figure that the model successfully identifies entities related to sustainability, such as \textit{'sustainability'}, \textit{'recycling'}, \textit{'clean technology'}, \textit{'business model'}, \textit{'repurposing'}, and even links corporations such as \textit{'Samsung'} to these entities. We can notice that multiple entities are interlinked in a meaningful way. 

\subsection{ChatGPT}

\begin{figure}[htbp]
\label{fig:fig_gpt}
\centerline{\includegraphics[width=0.4\textwidth]{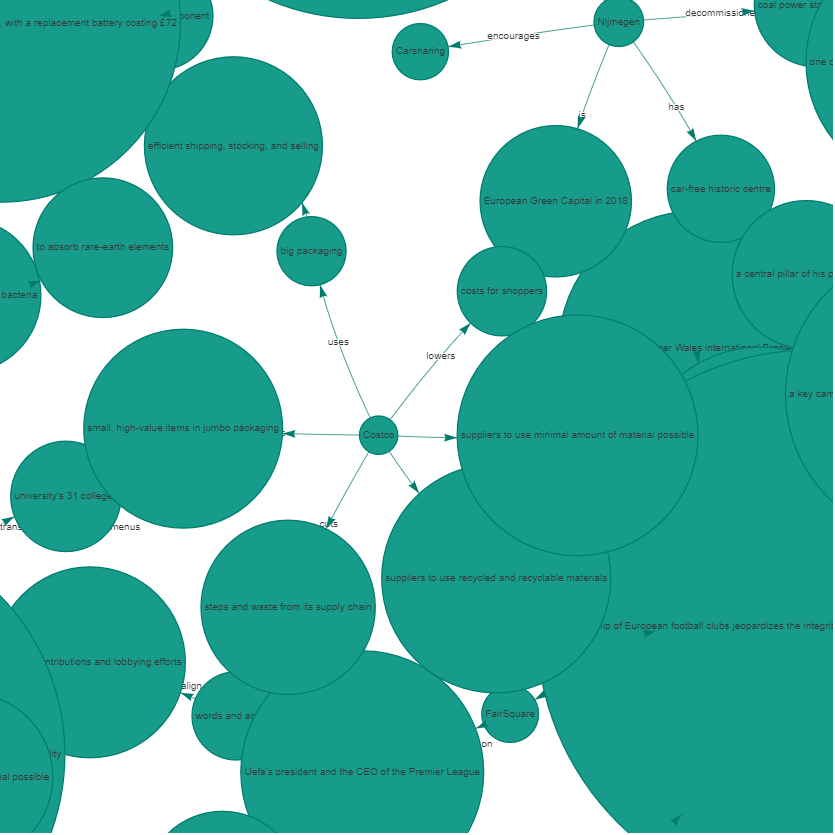}}
\caption{Subset of the Knowledge Base generated using the first experiment with ChatGPT. The Knowledge Base is displayed in a graph format where entities are represented as nodes and relations are represented as edges.}
\end{figure}

The same visualization for the Knowledge Base generated by the first experiment with ChatGPT is represented in this subsection. Fig. \ref{fig:fig_gpt} displays a subset of the extracted Knowledge Base.

We can see from the figure that ChatGPT is able to identify entities related to sustainability, but they are represented as phrases instead of concepts. For example, ChatGPT extracts \textit{'small high-value items in jumbo packaging'}, \textit{'steps and waste from its supply chain'}, and \textit{'suppliers to use recycled and recyclable materials'}, as entities. 

Although these phrases are related to sustainability, they do not represent specific entities. This happens as a result of the fact that ChatGPT is a conversational model trained on a task to generate responses to a provided prompt and not specifically trained to be able to recognize entities and relations. On the other hand, ChatGPT is able to identify some concepts that REBEL does not, and additionally, it is able to link corporations to specific sustainability-related phrases.

Prompt engineering \cite{Saravia_Prompt_Engineering_Guide_2022} is of great importance when it comes to the results generated from ChatGPT \cite{openai_gpt-4_2023}. Since it is a generative model, small variations in the input sequence can create large differences in the produced output. 

Observing the full Knowledge Base generated using ChatGPT, most of the time, the extracted entities represent phrases or whole sentences, which is not beneficial for creating a Knowledge Base because it's hard to normalize the entities and relations and create a more general ontology consisting of the concepts represented in the graph.

For this reason, we conducted the second experiment with ChatGPT, where we defined a more detailed prompt and instructed ChatGPT to generate an ontology based on each article it sees and additionally define instances of the generated ontology based on the information present in each article. 

\begin{figure}[htbp]
\label{fig:fig_chatgpt_article_1}
\centerline{\includegraphics[width=0.9\textwidth]{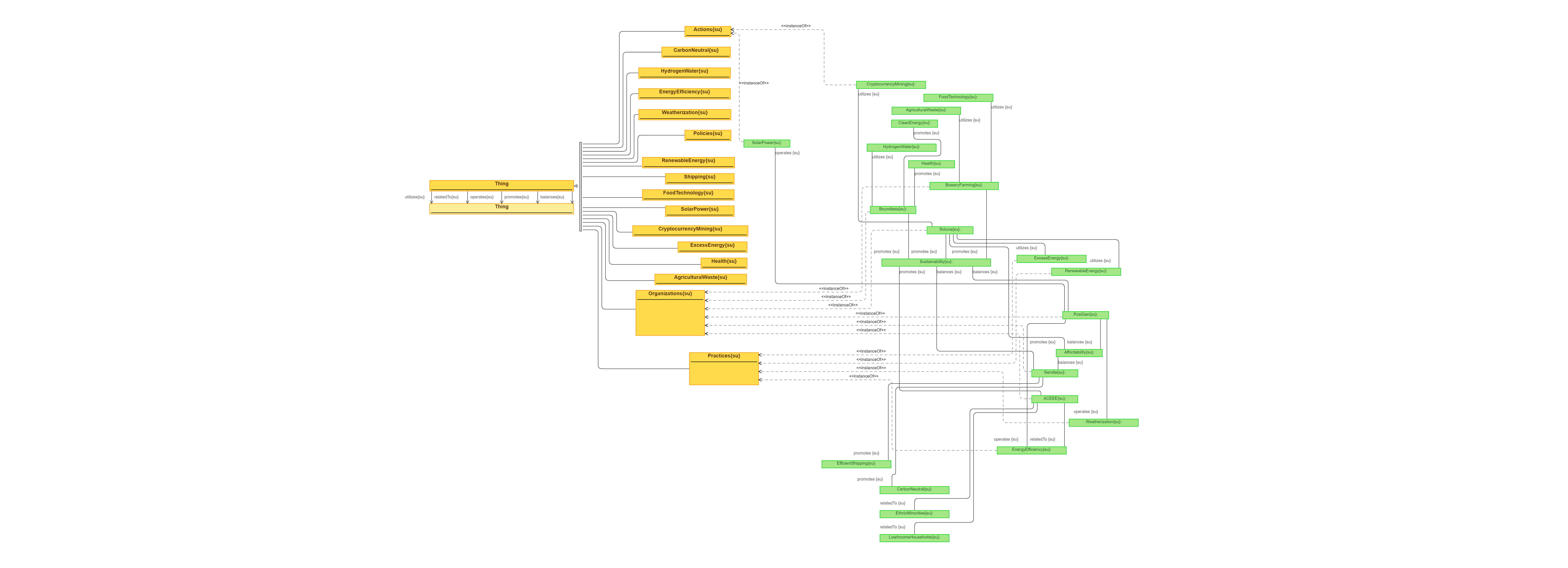}}
\caption{Knowledge Base generated with ChatGPT for the first article. The identified concepts are represented as yellow rectangles, and the instances are represented with green rectangles.}
\end{figure}

Figure \ref{fig:fig_chatgpt_article_1} presents the results of the refined prompt, with the ontology and instances generated from a single article out of the 94 collected articles.

Not only does ChatGPT create an ontology using the concepts it was instructed to use, but it also defines classes on its own and is able to create instances of most of the classes accurately.

As an example, it identifies the entity \textit{"Soluna"} as an \textit{"instanceOf"} the class \textit{"Organizations"}. Furthermore, it is able to identify the triplet $<$Soluna, utilizes, Excess Energy$>$, and $<$Excess Energy, instanceOf, Practices$>$. 

These types of triplets already start representing an initial knowledge base,  which can answer queries on companies that implement practices that use excess energy. Although the hierarchy of concepts can be better defined so that more complex queries can be answered, this method represents a solid start in building a shared Knowledge Base, using only unstructured texts.

\begin{figure}[htbp]
\label{fig:fig_chatgpt_article_2}
\centerline{\includegraphics[width=0.5\textwidth]{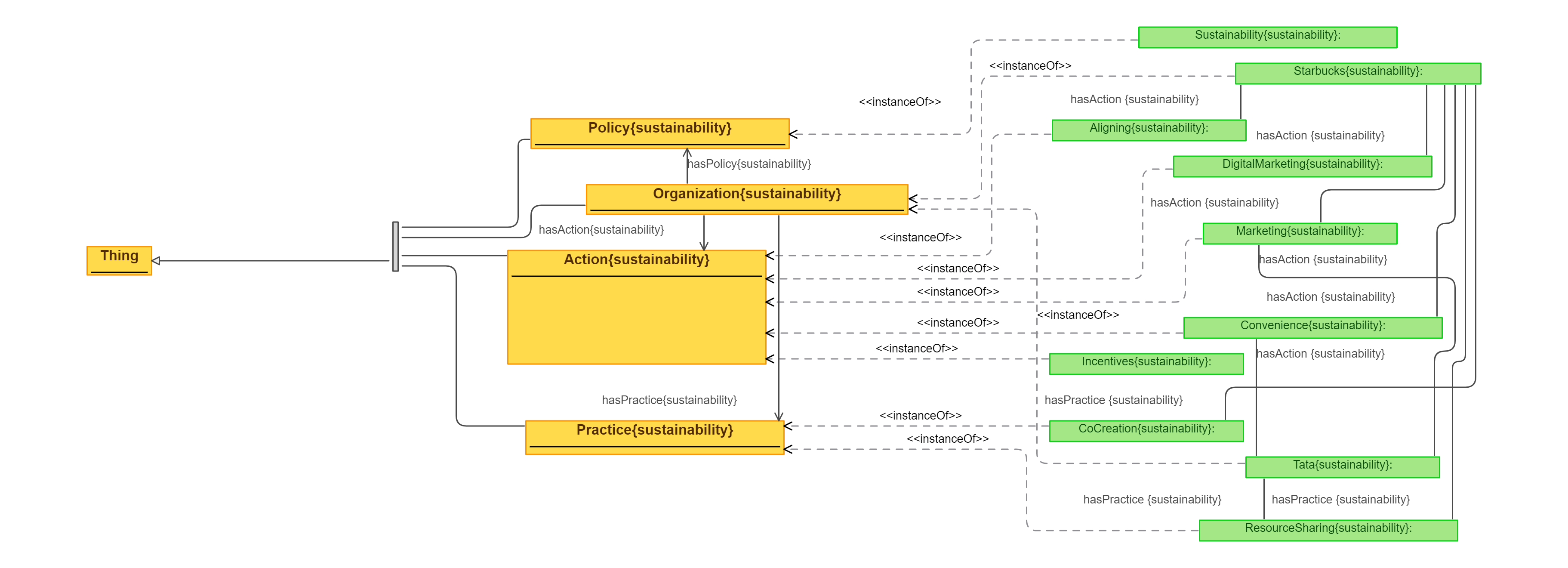}}
\caption{Knowledge Base generated with ChatGPT for the second article. The identified concepts are represented as yellow rectangles, and the instances are represented with green rectangles.}
\end{figure}

Using another article, the ontology and instances given in Fig.\ref{fig:fig_chatgpt_article_2} have been generated. Looking at this second example, we can see that ChatGPT links practices, actions, and policies to the organizations, which was not the case in the previous example.

Additionally, it identifies the triplets $<$Starbucks, instanceOf, Organization$>$, and $<$Starbucks, hasPractice, ResourceSharing$>$. This also allows for answering complex queries in the sustainability domain.

While the consistency of the generated ontologies may be limited, our analysis reveals that there are significant similarities between them. Therefore, future research can explore methods for unifying these ontologies across all articles, which has the potential to enhance the overall definition of concepts and their interrelationships in the sustainability domain.

It is important to mention that due to the limitations of the length of the input prompt passed to ChatGPT, it was not possible to prompt the model first to define an ontology based on all articles on sustainability and then create instances from all the other articles using the same ontology.

\subsection{Quality Evaluation}

Since the evaluation of a Knowledge Base cannot be created in an automated way based on some metric, when ground truth data is not available, we need to utilize qualitative principles in order to evaluate the results. Based on the practical framework defined in the study \cite{chen2019practical}, the following 18 principles identified:

\begin{enumerate}
    \item Triples should be concise
    \item Contextual information of entities should be captured
    \item Knowledge graph does not contain redundant triples
    \item Knowledge graph can be updated dynamically 
    \item Entities should be densely connected
    \item Relations among different types of entities should be included 
    \item Data source should be multi-field 
    \item Data for constructing a knowledge graph should in different types and from different resources
    \item Synonyms should be mapped, and ambiguities should be eliminated to ensure reconcilable expressions
    \item Knowledge graph should be organized in structured triples for easily processed by machine 
    \item The scalability with respect to the KG size 
    \item The attributes of the entities should not be missed 
    \item Knowledge graph should be publicly available and proprietary 
    \item Knowledge graph should be an authority 
    \item Knowledge graph should be concentrated
    \item The triples should not contradict each other 
    \item For domain-specific tasks, the knowledge graph should be related to that field 
    \item Knowledge graph should contain the latest resources to guarantee freshness 
\end{enumerate}

According to these principles, in our use case, we manually inspected the Knowledge Graphs generated with the proposed methods, and we can conclude that the second ChatGPT approach creates a Knowledge Graph of greater quality compared to the other two Knowledge Bases. 

However, it should be noted that to create these Knowledge Bases, a few steps of refining the answers from ChatGPT are needed. Sometimes the produced output is erroneous and needs to be corrected before proceeding. Thus, this calls for methods for automatically identifying incorrect OWL syntax and requesting to fix the previous output.

\section{Conclusion}
In this paper, we presented a Natural Language Processing-based method for constructing a Knowledge Graph on the topic of sustainability using raw documents available on the Web. The study demonstrated that meaningful information could be extracted from unstructured data through an automated process, which can subsequently be utilized for decision-making and process modeling. The focus on sustainability served as a concrete use case, illustrating the effectiveness and potential of the presented approach.

Although the experiments were conducted on the use case of sustainability, the primary emphasis is on the methodology itself, which lays the foundation for empirical analysis of qualitative data derived from various sources. The construction of a Knowledge Base using the presented approach can serve as a first step for analyzing diverse aspects of any subject matter and answering complex queries based on the gathered information.

In future research, first, we plan to adopt a more formal framework for assessing the quality of generated knowledge graphs. Such a framework will enable us to effectively evaluate the quality of KGs and provide a standardized means of assessing their overall quality. We also want to extend the presented methodology to other domains, unifying generated knowledge bases and employing graph-based modeling to predict missing links between concepts and relationships for a given domain.
\bibliographystyle{ieeetr}
\bibliography{conference_101719}

\end{document}